# Unsupervised Classification Using Immune Algorithm

M.T. Al-Muallim

Department of Computer Engineering & Automation,
Faculty of Mechanical and Electrical Engineering,
Damascus University, Syria

R. El-Kouatly

Department of Networking and systems,
Faculty of Information technology Engineering
Damascus, Syria

## ABSTRACT

Unsupervised classification algorithm based on clonal selection principle named *Unsupervised Clonal Selection Classification* (UCSC) is proposed in this paper. The new proposed algorithm is data driven and self-adaptive, it adjusts its parameters to the data to make the classification operation as fast as possible. The performance of UCSC is evaluated by comparing it with the well known *K*-means algorithm using several artificial and real-life data sets. The experiments show that the proposed UCSC algorithm is more reliable and has high classification precision comparing to traditional classification methods such as *K*-means.

## General Terms

Pattern Recognition, Algorithms.

## Keywords

Artificial Immune Systems, Clonal Selection Algorithms, Clustering, *K*-means Algorithm.

## 1. INTRODUCTION

Unsupervised classification which also known as data clustering is defined as the problem of classifying a collection of objects into a set of natural clusters without any a priori knowledge. Many clustering methods were proposed. These methods can be basically classified into two categories: hierarchical and partitional. In contrast to hierarchical clustering, which yields a successive level of clusters by iterative fusions or divisions, partitional clustering assigns a set of data points into *K* clusters without any hierarchical structure. This process usually accompanies the optimization of a criterion function [1]. One of the widely used partitional method is the *K*-means algorithm. It is an iterative hill-climbing algorithm. Many of stochastic optimization methods were used in clustering problem Such as genetic algorithms GAs [2-5], simulated annealing (SA) [6-8] and tabu search (TS) [9].

Artificial Immune Systems (AIS) is a field of study devoted to the development of computational models based on the principles of the biological immune system. It is an emerging area that explores and employs different immunological mechanisms to solve computational problems [10]. A lot of immune algorithms were developed aiming to finding solutions to a broad class of complex problems. Applications of AIS have included the following areas: clustering and classification [11-15], anomaly detection [16-18], optimization [13, 19], control [20-21], computer security, learning, bio-informatics, image processing, robotics, virus detection and web mining [22].

Many of the immune algorithms use principles inspired by the clonal selection theory of acquired immunity. The clonal selection principle is used by the immune system to describe the basic features of an immune response to an antigenic stimulus. It establishes the idea that only those cells that recognize the antigens proliferate, thus being selected against those which do not. The process of proliferating called clonal expansion. The selected cells are subject to an affinity maturation process which improves their affinity to the selective antigens [23]. The first clonal selection algorithm was proposed by [13] which named CLONALG and used for optimization. Other versions of clonal selection algorithm are designed to improve the performance of CLONALG such as [24-26].

In this study, we tried to investigate the possibility of using clonal selection algorithm as stochastic optimization methods for clustering. Unsupervised classification algorithm based on clonal selection principle named *Unsupervised Clonal Selection Classification* (UCSC) is proposed. The new algorithm is data driven and self-adaptive, it adjusts its parameters to the data to make the classification operation as fast as possible. The proposed approach has been tested on several artificial and real-life data sets and its performance is compared with the well known *K*-means algorithm [1]. Experiments show that UCSC algorithm is more reliable and has high classification precision comparing to traditional classification methods such as *K*-means.

## 2. UNSUPERVISED CLONAL SELECTION CLASSIFICATION (UCSC)

### 2.1 Basic Principle

In UCSC, clustering problem is considered as optimization problem and the objective is to find the optimal partitions of data where the resulting clusters tend to be compact as possible. A simple criterion which is the within cluster spread is used in UCSC, this criterion needs to be minimized for good clustering. Unlike *K*-means which uses the square-error criterion to measure the within cluster spread, UCSC uses the sum of the Euclidean Distances of the points from their respective cluster centroids as clustering metric and uses clonal selection algorithm as clustering algorithm which ensures finding the global optima when most of others algorithms such as *K*-means stuck into local optima. The number of clusters *K* is supposed to be known and the appropriate cluster centers $m_1$, $m_2,…,m_k$ have to be found such that the clustering metric *J* is minimized. Mathematically, the clustering metric $J$ for the $K$ clusters $C=\{C_1, C_2, …, C_K\}$ is given by the following equation:

$$J(\Gamma, \mathbf{M}) = \sum_{i=1}^{K} \sum_{j=1}^{N} \gamma_{ij} \left\| x_j - m_i \right\| \tag{1}$$

where $x_j \in \Re^d$, $j = 1,…, N$ are data points, $\Gamma = \{\gamma_{ij}\}$ is a partition matrix witch given by the eq. (2), M is centroid matrix witch given by the eq. (3) and $m_i \in \Re^d$, $i = 1,…, K$ is the mean for the Ci cluster with Ni data points.

$$\gamma_{ij} = \begin{cases} 1 & if \quad x_j \in C_i \\ 0 & oterwise \end{cases} \text{ with } \sum_{i=1}^{K} \gamma_{ij} = 1 \;\; \forall j \tag{2}$$

$$\mathbf{M} = [m_1, m_2, …, m_K] \quad \text{where} \quad m_i = \frac{1}{N_i} \sum_{j=1}^{N} \gamma_{ij} x_j \;, \; i = 1,…, K \tag{3}$$

The task of clonal selection algorithm is to search for the appropriate cluster centers wherefore $J$ is minimized. Based on clustering criterion, UCSC is supposed to give right results if the clusters are compact and hyperspherical in shape.

## 2.2 Clonal Selection Algorithm

All clonal selection algorithms have the same basic steps which are sumerized as follows:
*generate population P of antibodies (candidate solutions)*
*while stopping criterion is not met do {*
   *clone P based on their affinity*
   *Submit the result population to hypermutation scheme*
   *select the highest affinity solution to form new population*
*maintain diversity in the population }*
*select the highest affinity antibody to form the immune memory*
which is the solution to the problem.

There are certain critical issues that must be taken into consideration while designing and running a clonal selection algorithm such as representing the solution, maintaining diversity in population, affinity metric and hypermutation mechanism. Even a small change in any of these aspects may lead to a considerable change in the performance of clonal selection algorithms [26].

The UCSC algorithm is summarized as follows:
*Initialization: generate population P of n antibodies (candidate solutions) randomly*
*For every generation do: {*
   *affinity measure of all antibodies in P*
   *clone P generating a population PC*
   *submit PC to hypermutation scheme generating Pm*
   *consolidate P & Pm*
   *affinity measure of all antibodies*
   *re-select the n highest affinity to form P*
   *Generate new L individuals (randomly)*
   *replace the L lowest affinity antibodies in P with the new ones }*
*finally: select the highest affinity antibody in P which is the solution*
These steps will be described in details next.

## 2.3 Solution Representing

Each antibody in P forms a string of real numbers representing the K cluster centers.

For $d$-dimensional space, the length of the string is $d*K$ number, where the coordinates of the centers are localized in sequence.

$$p = [Ab_1, Ab_2, ……Ab_n] \tag{4}$$
$$Ab_l = [m_{11}, m_{12}, …m_{1d}, m_{21}, ……m_{Kd}] \;, \; l = 1,…, n \tag{5}$$

The first $d$ numbers represent the $d$ dimensions of the first cluster center; the next $d$ positions represent those of the second cluster center, and so on.

## 2.4 Affinity Metric

To measure the affinity of an antibody, the clusters are formed according to the centers encoded in the antibody under consideration, this is done by assigning each point $x_j \in \Re^d$, $j = 1,…, N$ to one of the clusters $C_i$ whose center are the closest to the point. After the clustering is done, the new cluster centroids are calculated by finding the mean points of the respective clusters, then clustering criterion $J$ is calculated by eq. (1). The affinity is defined as:

$$aff = \frac{1}{J} \tag{6}$$

The maximum value of the affinity standing for the minimum value of $J$. Zero is assigned to the affinity if any cluster becomes empty.

## 2.5 Cloning

Antibodies in $P$ will be cloned proportionally to their affinities, the higher the affinity the higher the number of clones generated for the antibody. The antibodies were sorted in descending order according to their affinity and then the amount of clones generated for the antibodies was given by:

$$nc_l = round\left( \frac{\beta \, n}{l} \right) \tag{7}$$

Where $nc_l$ is number of clones and $\beta$ is clonal factor.

## 2.6 Hypermutation Mechanism

Every antibody in $P_C$ is submitted to a mutation that is inversely proportional to the affinity and this is done according to the following equations:

$$Ab^* = Ab + \alpha \, N(0,1) \tag{8}$$
$$\alpha = \rho \, e^{-aff} \tag{9}$$

Where $Ab^*$ is the resulting antibody of mutate $Ab$, $N(0,1)$ is a matrix of $d*K$ Gaussian random variables with zero mean and standard deviation $\sigma=1$ , $aff$ is the affinity of the antibody, which is normalized in the range [0 1]. $\alpha$ is a factor that resizes the value of the Gaussian mutation and it is inversely proportional to the affinity, $\rho$ is a factor that controls the range of $\alpha$. To make the algorithm fast and data driven this factor is given as:

$$\rho = (\max_{data} - \min_{data}) \qquad (10)$$

Where $\max_{data}$ and $\min_{data}$ are the maximum and the minimum values of the data features at all dimensions. In this way, the mutation probability depends on the affinity of the antibody and also on the scope of search.

## 2.7 New Antibodies Generator

To generate new random solutions, the scope of search which is the data distribution range in the feature space was determined. The range of data was calculated using the upper and the lower limit of the data at every dimension:

$$UL_{data} = [UL_1, UL_2, \cdots, UL_d] \qquad (11)$$

$$LL_{data} = [LL_1, LL_2, \cdots, LL_d] \qquad (12)$$

Where $UL_{data}$, and $LL_{data}$ are matrixes of the upper and lower limits of the features respectively. Then new random solution is generated using:

$$Ab_{new} = LL_{data} + (\text{diag}((UL_{data} - LL_{data})^T \times rand))^T \qquad (13)$$

Where $rand$ is a matrix of $d*K$ random variables with uniform probability distribution within the range [0 1]. This random solutions generator is used to insure a fast and accurate performance of UCSC and accelerate the convergence rate of the algorithm since all solutions are in the scope of search.

## 3. EXPERIMENTS

The UCSC was tested using several artificial and real-life data sets, then compared with the well known $K$-means algorithm [1]. The UCSC was tested with the following parameters $n$=10, $\beta$ =5, $L$=4, and number of generations $gen$=30. For $K$-means algorithm [1] 1000 as a maximum number of iterations was used in case it does not terminate normally. At every experiment the algorithms were run for 100 times with different random initial configurations To provide statistical evaluation of the performance.

The data sets are described below:

## 3.1 Artificial Data Sets

*Dataset* 1: An artificial dataset consisting of overlapping two classes (100 patterns each) with bivariate Gaussian density with the following parameters:

$m_1$=(0.1,  0.1),  $m_2$=(0.35, 0.1),

$$\sum\nolimits_1 = \sum\nolimits_2 = \begin{bmatrix} 0.11 & 0 \\ 0 & 0.1 \end{bmatrix}, \text{ where } \Sigma \text{ is covariance matrix.}$$

The dataset is shown in Figure (1).

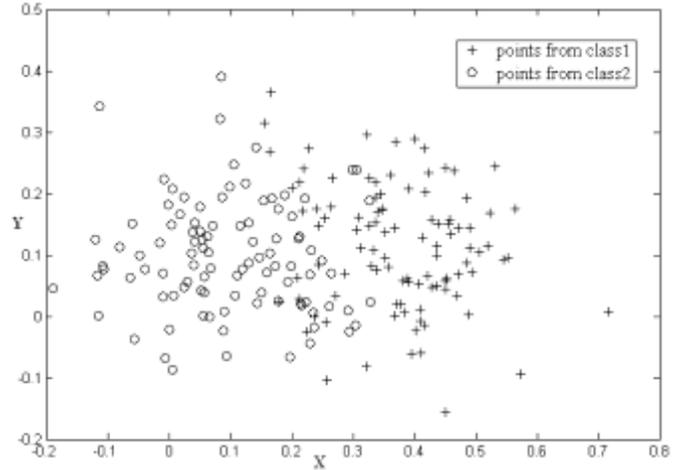

**Figure .1: An artificial dataset 1 with two classes.**

*Dataset* 2: An artificial dataset consisting of nine classes (25 patterns each) with bivariate Gaussian density with the following parameters:

$m_1$=(0.1,  0.1),  $m_2$=(0.1,  0.5),  $m_3$=(0.1,  0.9),  $m_4$=(0.5,  0.1), $m_5$=(0.5,  0.5),  $m_6$=(0.5,  0.9),  $m_7$=(0.9,  0.1),  $m_8$=(0.9,  0.5), $m_9$=(0.9, 0.9),

$$\sum\nolimits_1 = \sum\nolimits_2 = \sum\nolimits_3 = \ldots = \sum\nolimits_9 = \begin{bmatrix} 0.08 & 0 \\ 0 & 0.08 \end{bmatrix}$$

The dataset is shown in Figure (2).

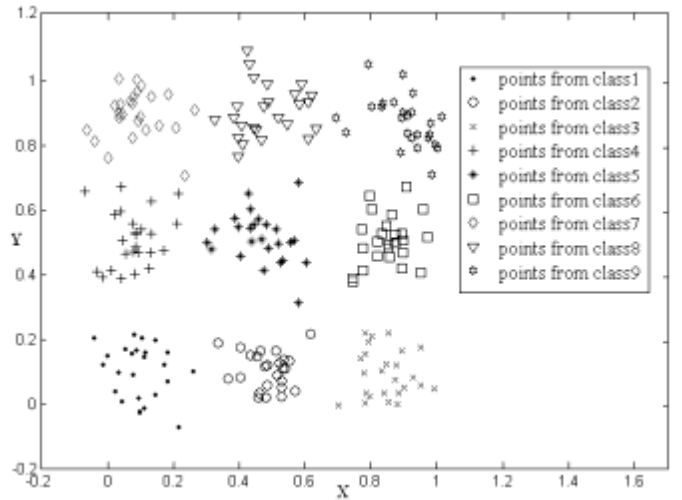

**Figure .2: An artificial dataset 2 with nine classes**

*Dataset* 3: An artificial dataset consisting of three classes (50 patterns each) with tripartite Gaussian density with the following parameters:

$m_1$=(1, 1, 1), $m_2$=(2, 2.5, 2.5), $m_3$=(2, 3, 3),

$$\sum{}_1 = \sum{}_2 = \sum{}_3 = \begin{bmatrix} 0.3 & 0 & 0 \\ 0 & 0.3 & 0 \\ 0 & 0 & 0.3 \end{bmatrix}$$

The dataset is shown in Figure (3).

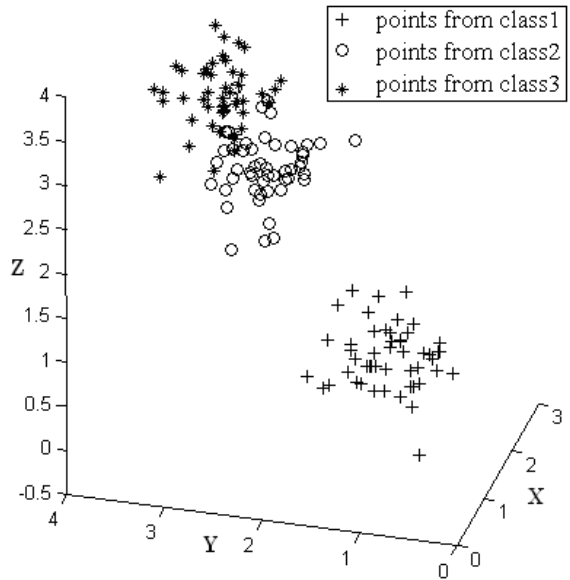

**Figure .3: An artificial dataset 3 with three classes.**

### 3.2 Real-life Data Sets

The following real-life dataset has being tested:

1- *Iris dataset* [27] consists of 150 four-dimensional patterns in three classes (50 patterns each) represent different categories of iris flowers which have four feature values. The four feature values represent the sepal length, sepal width, petal length and the petal width in centimeters. The three classes are: Iris Setosa, Iris Versicolor and Iris Virginica.

2- *Wisconsin Breast Cancer dataset* [28] consists of 699 nine-dimensional patterns in two classes which are Benign (458 patterns) and Malignant (241 patterns). The nine features are: Clump Thickness, Uniformity of Cell Size, Uniformity of Cell Shape, Marginal Adhesion, Single Epithelial Cell Size, Bare Nuclei, Bland Chromatin, Normal Nucleoli and Mitoses.

## 4. RESULTS

The best classification results of UCSC and *K*-means after run for 100 times are shown in Table (1) which includes the obtained classification accuracy for all datasets. As we can see from Table (1) the UCSC algorithm provides better accuracy compared with *K*-means algorithm.

**Table 1: classification result for all datasets**

| Dataset | UCSC | *K*-means |
|---|---|---|
| Dataset 1 | 88% | 86% |
| Dataset 2 | 97.78% | 97.33% |
| Dataset 3 | 91.33% | 91.33% |
| Iris dataset | 90% | 89.33% |
| Breast Cancer dataset | 96.11% | 95.7% |

The experiments show that *K*-means algorithm got stuck at sub-optimal solutions even for simple data but UCSC did not exhibit any such behavior. Table (2) shows the best values of *J* and its percentages of the total runs of UCSC and *K*-means algorithms for every dataset. As we can see from Table (2) for all datasets, UCSC finds better solutions than *K*-means algorithm and the clusters formed by UCSC are more compact than those formed by *K*-means algorithm. The results show that UCSC algorithm is more reliable than *K*-means algorithm because it finds the best solution all the time unlike *K*-means which did not find the best solution all the times.

**Table 2: Values of *J* for the different datasets.**

| Dataset | UCSC | | *K*-means | |
|---|---|---|---|---|
| | *J* | percents | *J* | percents |
| Dataset 1 | 25.141 | 100% | 25.166 | 100% |
| Dataset 2 | 21.597 | 100% | 21.906 | 40% |
| Dataset 3 | 70.628 | 100% | 70.653 | 75% |
| Iris dataset | 97.101 | 100% | 97.205 | 80% |
| Breast Cancer dataset | 3048.2 | 100% | 3051.3 | 100% |

For all experiments UCSC found the solution in less than 30 generations

## 5. DISCUSSION

means algorithm using several artificial and real-life data sets. The experiments show that the proposed UCSC algorithm is more reliable because it finds the best solution all the time unlike *K*-means which got stuck at sub-optimal solutions. UCSC algorithm has high classification precision comparing to *K*-means algorithm.

The new proposed algorithm is data driven and self-adaptive, it adjusts its parameters to the data to make the classification operation as fast as possible.

UCSC algorithm has many advantages comparing to other evolutionary algorithms. One is the small population size *n*=10 where most of other evolutionary algorithms need at lest population size of 100. Second it found the solution in less than 30 generations.

Based on clustering criterion used in UCSC it supposed to give right results if the clusters are compact and hyperspherical in shape.

## 6. CONCLUSION

In this paper, unsupervised classification algorithm based on clonal selection principle named *Unsupervised Clonal Selection Classification* (UCSC) is designed to find the optimal partition between the data. It uses within cluster spread criterion as a clustering criterion. The criterion is based on Euclidean distance between the data in the clusters. The new algorithm is data driven and self-adaptive, it adjusts its parameters to the data to make the classification operation as fast as possible.

UCSC is tested on several artificial and real-life data sets and its performance is compared with the well known *K*-means algorithm [1]. The experiments show that UCSC algorithm has classification precision higher than *K*-means algorithm which got stuck at sub-optimal solutions even for simple data sets. The new algorithm finds the solution in thirty generations only and it uses a small population size $n=10$ where most of other evolutionary algorithms need at lest population size of 100. UCSC gives good results if the clusters are compact and hyperspherical in shape